\documentclass[times,twocolumn,final]{elsarticle}
\usepackage{framed,multirow}
\usepackage{adjustbox}
\usepackage{amssymb}
\usepackage{latexsym}
\usepackage{url}
\usepackage[hidelinks]{hyperref}
\hypersetup{
  colorlinks   = true, 
  urlcolor     = blue, 
  linkcolor    = blue, 
  citecolor   = red 
}
\usepackage{graphicx}
\usepackage{tabularx}
\usepackage{times}
\usepackage{epsfig}
\usepackage{amsmath,empheq}
\usepackage{amssymb}
\usepackage{amsfonts}
\usepackage{bm}
\usepackage{booktabs}
\usepackage{subfig}
\usepackage{xfrac}
\usepackage{threeparttable}
\usepackage{multirow}
\usepackage{rotating}
\usepackage{changepage}
\usepackage{makecell}
\usepackage{floatrow}
\usepackage{pifont}
\usepackage{subfig}
\usepackage{booktabs}
\newcolumntype{P}[1]{>{\centering\arraybackslash}p{#1}}
\usepackage{lineno}
\modulolinenumbers[5]

\journal{}
\usepackage{tabu}
\usepackage{xcolor}
\usepackage{soul}

\makeatletter

\newcommand*{\@rowstyle}{}

\newcommand*{\rowstyle}[1]{
  \gdef\@rowstyle{#1}%
  \@rowstyle\ignorespaces%
}

\newcolumntype{=}{
  >{\gdef\@rowstyle{}}%
}

\newcolumntype{+}{
  >{\@rowstyle}%
}

\makeatother
\begin{document}
\begin{frontmatter}
\title{Multi-task learning with cross-task consistency for improved depth estimation in colonoscopy}
\author[1]{Pedro Esteban Chavarrias Solano}
\author[1]{Andrew Bulpitt}
\author[2]{Venkataraman Subramanian}
\author[1]{Sharib {Ali}\corref{corr1}}\cortext[corr1]{Corresponding author}
\ead{s.s.ali@leeds.ac.uk}
\address[1]{School of Computing, Faculty of Engineering and Physical Sciences, University of Leeds, Leeds, LS2 9JT, United Kingdom}
\address[2]{Leeds Teaching Hospitals NHS Trust, Leeds, UK}
%
\begin{abstract}
Colonoscopy screening is the gold standard procedure for assessing abnormalities in the colon and rectum, such as ulcers and cancerous polyps. Measuring the abnormal mucosal area and its 3D reconstruction can help quantify the surveyed area and objectively evaluate disease burden. However, due to the complex topology of these organs and variable physical conditions, for example, lighting, large homogeneous texture, and image modality estimating distance from the camera (\textit{aka} depth) is highly challenging. Moreover, most colonoscopic video acquisition is monocular, making the depth estimation a non-trivial problem. While methods in computer vision for depth estimation have been proposed and advanced on natural scene datasets, the efficacy of these techniques has not been widely quantified on colonoscopy datasets. As the colonic mucosa has several low-texture regions that are not well pronounced, learning representations from an auxiliary task can improve salient feature extraction, allowing estimation of accurate camera depths. 
In this work, we propose to develop a novel multi-task learning (MTL) approach with a shared encoder and two decoders, namely a surface normal decoder and a depth estimator decoder. Our depth estimator incorporates attention mechanisms to enhance global context awareness. We leverage the surface normal prediction to improve geometric feature extraction. Also, we apply a cross-task consistency loss among the two geometrically related tasks, surface normal and camera depth. 
We demonstrate an improvement of 14.17\% on relative error and 10.4\% improvement on $\delta_{1}$ accuracy over the most accurate baseline state-of-the-art BTS approach. All experiments are conducted on a recently released C3VD dataset; thus, we provide a first benchmark of state-of-the-art methods.
\end{abstract}
\end{frontmatter}
\section{Introduction}
\label{sec:introduction}
Colorectal cancer (CRC) is among the third most common type of cancer in the world, imposing a healthcare burden globally. The estimated number of new CRC cases in 2023 will likely increase to 153,020~\citep{siegel2023}. Optical endoscopy is the gold standard procedure for diagnosing and treating CRC~\citep{rex2015}. Despite its great potential, the colonoscopic procedure is subject to the clinician's experience as they have to deal with a complex anatomical environment, imaging artefacts, and a limited field of view. A retrospective analysis of clinical endoscopic video realised that 9.6\% of the colon surface is never imaged during the screening procedure~\citep{mcgill2018}. Those missed regions can contribute to an estimated 22\% of precancerous undetected lesions found by~\cite{vanrijn2006}. The development of an intelligent system to reduce the missed detection rate and guide the clinician to potential regions of interest has caught the attention of the medical computer vision research community. Several methods have been developed to detect and segment polyp instances accurately. Unlike polyp detection and segmentation methods, 3D computer vision techniques in colonoscopy have not been widely explored. Some applications within this field cover lesion extent prediction~\citep{abdelrahim2022,ali2021}, observational coverage of the colon~\citep{armin2016,bobrow2022}, and 3D reconstruction~\citep{zhang2021}. 

The development of 3D computer vision applications in colonoscopy is limited due to the difficulty in acquiring ground truth labels. Compared to labelled datasets required for training detection and segmentation methods, acquiring datasets with accurate surface information (e.g., surface normal vectors and depth maps) for 3D scene understanding is far from practical during a clinical endoscopy procedure. Hence, the development of commercial and computed tomography-derived silicone models has been suggested as an alternative method for data acquisition. The C3VD dataset leverages a novel technique for generating high-fidelity silicone phantom models of the colon with various textures and colors~\citep{bobrow2022}. The phantom model and a 2D-3D video registration algorithm were used to generate a ground truth dataset with pixel-level registration to a known 3D model. Unlike digital phantom datasets~\citep{rau2019,zhang2021dataset}, silicone phantom models are closer to real-world clinical datasets and reflect more accurately to an actual colonoscopic procedure.

Depth estimation is crucial for understanding geometry structure and a fundamental task in computer vision for 3D scene reconstruction. Most depth estimation approaches initially relied on stereo matching and triangulation methods to calculate the disparity of two 2D images. However, these binocular-based depth estimation methods require at least two fixed cameras. In addition, capturing enough features to match between images becomes challenging when the scene does not have enough texture~\citep{ming2021}. Due to the spatial constraint imposed by the lumen (e.g., size and complex non-uniform shape) of the gastrointestinal tract, monocular systems have been more attractive than stereo systems. Monocular depth estimators aim to learn a mapping between a single RGB image and their corresponding depth values by capturing features that represent geometric structures. 
Single-view deep learning methods~\citep{lee2019, yuan2022, piccinelli2023} for monocular depth estimation make use of monocular visual cues, e.g., texture gradients and lighting variations. These methods learn to incorporate scene priors without the need to compute camera motion. While this is an advantage, they usually perform well only in similar samples to those presented during the training stage, making these networks harder to generalise. Their accuracy is also affected by the essential ambiguity of the problem as an infinite number of world scenes and camera positions could have produced a given image~\citep{eigen2014,lee2019,zhang2021,piccinelli2023}. Therefore, recent approaches suggest using multi-task learning schemes as they leverage auxiliary tasks with depth-related features~\citep{ming2021}. Adopting an additional task aims to enhance the extraction of relevant geometrical cues during the encoding stage, leading to a better performance overall~\citep{ming2021,chen2020,qi2018}. 
For example, joint learning of depth and optical flow~\citep{zou2018, chen2020} and joint learning of depth and surface normal~\citep{bae2022} have shown competitive performances on natural scene datasets. Multi-task learning approaches have been extensively studied in natural scenes, but their applicability in the colonoscopy domain has not been widely validated. In addition, consistency among tasks is desirable since both generate a particular domain representation of the same underlying reality~\citep{zou2018,zamir2020}. 


To this end, we propose Col3D-MTL, a novel multi-task learning with a cross-task consistency approach for joint monocular depth and surface normal prediction featuring attention mechanisms to improve global context awareness. We validate our study on a public colonoscopy dataset completely acquired using a silicone phantom model.
Our proposed framework can be summarised on the following three main contributions: 
\begin{enumerate}
    \item We propose a multi-task learning network with one shared encoder and two independent decoders to predict depth and surface normal maps, leading to an enhanced feature representation of the scene. We introduce novel unit normal computation blocks (UNC blocks) in our surface normal decoder that enable the accurate recovery of the geometrical orientation of the scene.
    \item We incorporate a weighted cross-task consistency loss between our predicted surface normal and a warped surface normal computed from our depth prediction using depth image gradients to enforce consistency among both geometrically related tasks explicitly.
    \item We set a new benchmark comprising state-of-the-art monocular depth estimation methods and validate our multi-task learning with cross-task consistency framework on a public colonoscopy dataset with pixel-wise ground truth labelled data acquired from a silicone phantom model, the C3VD dataset~\citep{bobrow2022}.
\end{enumerate}

The rest of the paper is organised as follows. Section~\ref{sec:related_work} presents state-of-the-art methods on monocular depth estimation, multi-task learning, and cross-task consistency. In Section~\ref{sec:materials_and_method}, we introduce the C3VD dataset used in this work and our proposed framework. Section~\ref{sec:resultsandExperiments} describes the training and ablation study setups followed in this work. We also present the evaluation metrics and the corresponding quantitative and qualitative results. In Section~\ref{sec:discussion}, we discuss the findings of our approach. Finally, our conclusions are presented in Section~\ref{sec:conclusion}.

\section{Related work}
\label{sec:related_work}
This section introduces the most relevant technical aspects needed to understand our contribution. The structure of this section starts with a review of related works to monocular depth estimation in computer vision and endoscopy, followed by a discussion about multi-task learning approaches found in the literature covering the natural scene and endoscopy domains. Finally, we describe the cross-task consistency methods.

\subsection{Monocular depth estimation in computer vision}
Unlike most traditional stereo matching and triangulation approaches, monocular depth estimation methods only require a single camera to generate a depth map. Even though promising, it is still an ill-posed problem to regress depth from a single image~\citep{ming2021}. 
The success of deep learning in many computer vision tasks was also translated into the monocular depth estimation task. Learning-based approaches for monocular depth estimation were first introduced in 2014 by~\cite{eigen2014}. Their proposed method consists of a convolutional coarse-scale network to predict depth at a global level, followed by a fine-scale network to incorporate finer details, such as object edges. 

In~\citep{lee2019}, an atrous spatial pyramid pooling (ASPP) module is used to leverage global context information, while the decoder applies local planar guidance at different resolutions to provide geometric guidance to the full-resolution depth map.~\cite{kim2020} propose a convolution-based encoder-decoder scheme with attention mechanisms embedded in their skip connections to generate refined multi-scale features. A global context module is also introduced at the network's bottleneck to capture representative features on a global scale. 
~\cite{patil2022} exploit the high degree of regularity in 3D scenes by using a piecewise planarity prior to leverage information from co-planar pixels to improve depth estimation.

Leveraging the enhanced global context understanding of transformer-based architectures,~\cite{bhat2021} propose an adaptive bin-width estimator based on a mini vision transformer (mViT) network~\citep{miniViT}. The idea behind this approach is to divide the depth range into several adaptive-width bins and predict the final depth map as a linear combination of the bin centres. All these methods have achieved state-of-the-art performance on popular natural scene depth prediction datasets (such as KITTI~\citep{geiger2012}, NYU~\citep{silberman2012}).
\subsubsection{Monocular depth estimation applied to endoscopy}
In contrast to natural scenes, estimating depth from endoscopy data is highly affected by the lack of ground truth labelled data, low texture, variable lighting conditions and the presence of artefacts, e.g., specularities, saturation, and blurring effects. Conditional generative adversarial networks (cGANs), e.g., pix2pix~\citep{isola2017}, have been used to estimate depth from monocular endoscopic images~\citep{rau2019, cheng2021}. However, one major drawback of cGANs is the lack of realistic detail and texture in their representations.

~\cite{mahmood2018} applied continuous conditional random fields (CRFs) and a convolutional neural network (CNN) to estimate depth from endoscopy images. However, one limitation of this approach is the generation of artefacts due to specular reflections.~\cite{yang2023} proposed a geometry-aware monocular depth estimation network based on ManyDepth~\citep{watson2021} and leverage a depth, smoothness, gradient, normal and geometric consistency losses to enhance depth predictions on endoscopy images. However, their normal loss only relies on the normal map generated from the predicted depth map, which does not faithfully represent the characteristics of the scene~\citep{bae2022}.

\subsection{Multi-task learning in computer vision}
The complementarity between depth and other geometrically-related features has recently been explored by computer vision researchers~\citep{qi2018,chen2020,long2021}. According to a survey on monocular depth estimation~\citep{ming2021}, many approaches suggested incorporating joint multi-task training, in which the extracted features between tasks were projected from one to the other for improved performance~\citep{zou2018, ma2021, bae2022}. For example,~\cite{chen2020} developed an architecture composed of two tightly coupled encoder-decoder networks to predict depth map and optical flow as primary and auxiliary tasks, respectively. The authors also introduced exchange blocks to effectively communicate between depth and optical flow networks and an epipolar layer that confines feature matching along the epipolar line. In GeoNet,~\cite{qi2018} proposed jointly predicting the depth and surface normal maps from a single image. The method uses two stream-CNNs (ResNet-50~\citep{he2016} and VGG-16~\citep{simonyan2015}) to predict the initial depth and surface normal maps. It then applies the depth-to-normal and normal-to-depth modules to refine surface normal and depth maps. However, the depth-to-normal network solves a pre-trained least square equation from the initial depth map followed by a residual module to enhance the final prediction, which is not learned in an end-to-end fashion. Similarly, normal-to-depth also solves linear equations through a kernel regression module to infer depth from surface normal. 

\subsubsection{Multi-task learning applied to medical image analysis}
Multi-task learning approaches have been studied for breast cancer segmentation and classification~\citep{wang2023}, left ventricle quantification~\citep{xue2018}, and CT-based identification and quantification~\citep{goncharov2021}.~\cite{islam2021} proposed a spatio-temporal multi-task learning network with one shared-encoder and two spatio-temporal independent decoders for instrument segmentation and saliency on a robotic instrument segmentation dataset for endoscopy.
Other multi-task learning approaches on endoscopy have focused on monocular depth and motion estimation~\citep{shao2022,liu2022,recasens2021}.
A 3D colon reconstruction approach suggested by~\cite{ma2021} incorporates a multi-task recurrent neural network (RNN) that estimates depth and camera pose~\citep{wang2019} to improve the performance of a standard simultaneous localisation and mapping (SLAM) method.~\cite{zhang2021colde} leverage surface normal estimation to enhance feature extraction and improve their depth estimation performance. The authors used a shared encoder and two independent decoders for depth and surface normal prediction. However, both decoders have almost the same architecture, with the number of output channels only varying. In this work, we propose task-specific architectures for each decoder.

\subsection{Cross-task consistency loss}
In visual perception, different domain representations of the same underlying reality or scene are not independent, i.e., a consistent factor between them should exist. A general fully computational method for augmenting training is proposed in~\citep{zamir2020}. In this work, the authors introduce a loss for predicting domain $y_{1}$ from an input image, $x$, while imposing consistency with domain $y_{2}$. This approach compares prediction $y_{2}$ with the warped prediction of $y_{1}$ to domain $y_{2}$.
An unsupervised framework leveraging geometric consistency for training single-view depth and optical flow networks on an unlabeled dataset was proposed in~\citep{zou2018}. To enforce geometric consistency, the authors introduced a cross-task consistency loss to minimise the discrepancy between the estimated optical flow and a synthesised flow computed from the predicted depth map and an estimated 6D camera pose.

\section{Materials and Method}
\label{sec:materials_and_method}
In this section, we describe the dataset used in our work, and we also present the details of our proposed multi-task learning with cross-task consistency framework.

\subsection{The C3VD dataset}
In this study, we used the new publicly available Colonoscopy 3D Video Dataset (C3VD)~\citep{bobrow2022}, which is the first video dataset containing 3D pixel-wise ground truth labelled data entirely recorded with a high-definition (HD) clinical colonoscope. The authors created a complete 3D phantom model of the colon, which was digitally sculpted by a board-certified anaplastologist. A 3D-printed phantom model was generated and coated with silicone, silicone pigments and silicone lubricants to mimic the specular appearance of the mucosa, tissue features and vascular patterns.
The data acquisition was performed by mounting the tip of a colonoscope to the end-effector of a robotic arm with previously defined moving trajectories. Pixel-level ground truth for each frame was generated by moving a virtual camera along the digital model following the recorded trajectory of the robotic arm while rendering ground truth frames of the 3D model. 

\begin{figure}[!t]
    \centering
    \includegraphics[width=\textwidth]{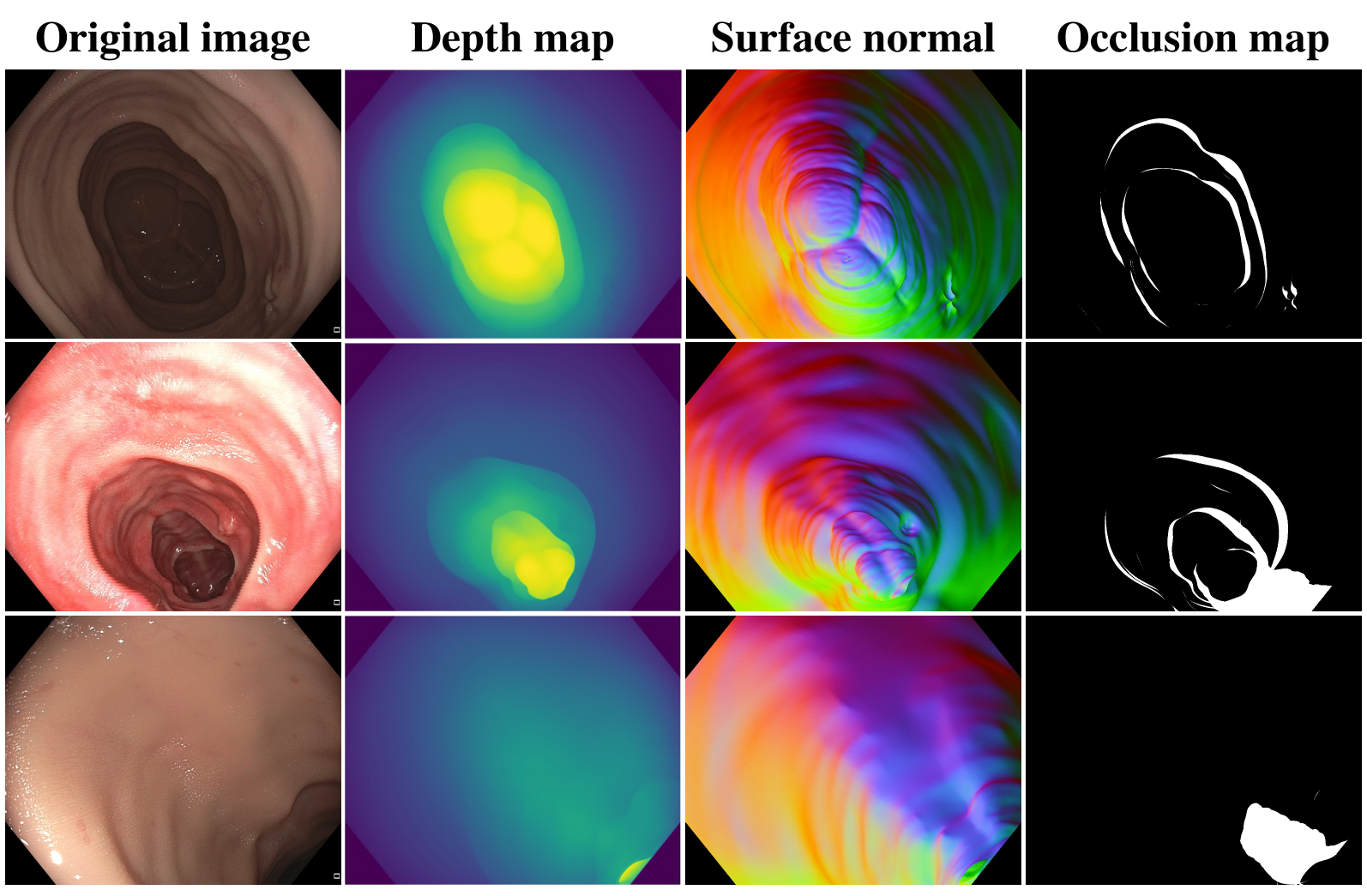}
    \caption{The C3VD dataset. Sample data including the original RGB image with its corresponding ground truth depth map, surface normal, and occlusion map~\citep{bobrow2022}.}
    \label{fig:C3VD}
\end{figure}

The dataset contains 10,015 frames with paired ground truth depth, surface normal, optical flow, occlusions, six-degrees-of-freedom (DoF) poses, coverage maps, and 3D models. The image resolution of all available data is $1080 \times 1350$ pixels.
In total, 22 videos covering four different colon segments (caecum, transverse, descending, and sigmoid), four texture variations, and three predefined trajectories are publicly available. Fig.~\ref{fig:C3VD} shows sample images with their corresponding ground truth depth map, surface normal, and occlusion map.

The train, validation and test splits used in this work are detailed in Table~\ref{table:dataset_split}. During the training stage, we followed a video-wise split in which we provided data from three colon segments (caecum, transverse, and sigmoid). Our models are validated on data collected from the same colon segments. For testing, we also include data from the descending colon segment, which was not given during the training stage.

\subsection{Method}
This subsection presents our proposed multi-task learning with a cross-task consistency framework for improved colonoscopy depth estimation. We describe the baseline depth estimation network selected in this study, followed by a review of our multi-task learning scheme and the cross-task consistency approach we incorporate into the network.
\renewcommand{\arraystretch}{1.3}
\begin{table}[t!]
\centering
\caption{Dataset split. Our dataset split follows a video-wise split. Each label corresponds to the colon segment, followed by the texture style and the predefined video trajectory. Here `c' refers to caecum, `s' refers to sigmoid, `t' refers to transverse and `d' for descending colon.}
\label{table:dataset_split}
\resizebox{\textwidth}{!}{
\begin{tabular}{l|l|c}
\toprule
\textbf{Split} & \multicolumn{1}{c|}{\textbf{Colonoscopy videos}} & \textbf{No. of frames} \\ \hline \hline
Training   & c1v1, c2v2, c2v3, c3v2, s1v3, s2v1, t1v1, & 6344\\
           & t1v3, t2v1, t2v2, t3v2, t3v1, t4v3                    \\ 
Validation & c4v2, c4v3,  s3v1, t2v3                         & 1738\\ 
Testing    & c2v1, d4v2, s3v2, t3v3                          & 1268\\
\bottomrule
\end{tabular} }
\end{table}
\subsubsection{Depth estimation network}
Our framework is based on the monocular depth estimation network proposed by~\cite{lee2019}. This method follows an encoder-decoder scheme, in which the encoder performs dense feature extraction while the decoder aims to regress the depth values. The decoder leverages local planar guidance (LPG) layers at different resolutions to improve the final output. The architecture of this network can be identified at the bottom of Fig.~\ref{fig:multitask_framework}.

The network uses ResNet-50~\citep{he2016} as its dense feature extractor, which outputs a feature map of $H/8$ resolution. The backbone is followed by an atrous spatial pyramid pooling module~\citep{chen2018} to extract contextual information at multiple dilation rates. During the decoding phase, internal outputs are recovered to their original resolution $H$ by a factor of 2 at each LPG block. The LPG block provides geometric guidance to the full-resolution depth map. A final $1\times 1$ convolutional layer is also used to extract the finest estimation ($\Tilde{c}^{1\times 1}$) after the last \textit{upconv} layer. All the estimated outputs ($\Tilde{c}^{k\times k}$) are concatenated and processed through a convolutional layer to compute the final depth estimation. The model is optimized by minimizing the scale-invariant logarithmic (SILog) error loss introduced by~\cite{eigen2014}.

\paragraph{Multi-scale local planar guidance (LPG block)}
Most monocular depth estimation networks following an encoder-decoder architecture just apply simple nearest neighbour up-sampling to recover the original resolution of the input image. Unlike those methods, LPG blocks guide features to the full resolution leveraging the local planar assumption~\citep{lee2019}. 
The LPG block consists of a stack of $1\times1$ reduction layers, which iteratively decrease the number of channels by a factor of two until it reaches a channel dimension of three. The resulting feature map ($H/k, H/k, 3$) is processed through two pathways to compute local plane coefficient estimations. The first pathway uses the first two channels to perform the conversion to unit normal vectors ($n_{1}, n_{2}, n_{3}$). A unit normal vector has only two degrees of freedom (DoF), polar and azimuthal angles from predefined axes. The two channels are regarded as polar $\theta$ and azimuthal $\phi$ angles and converted to unit normal vectors by Eq.~(\ref{eq:normal_vectors}).

\begin{empheq}[left=\empheqlbrace]{align}
 \label{eq:normal_vectors}
    & n_{1} = sin(\theta) cos(\phi) \nonumber\\
    &
    n_{2} = sin(\theta) sin(\phi)\\
    & n_{3} = cos(\theta)\nonumber
\end{empheq}

The second pathway estimates the perpendicular distance ($n_{4}$) between the plane and the origin. This pathway computes a sigmoid function from the third channel and multiplies its output with the maximum depth value. 
Finally, these 4D plane coefficients ($n_{1}, n_{2}, n_{3}, n_{4}$) are used to compute $k\times k$ local depth cues using the ray-plane intersection Eq.~(\ref{eq:ray_plane_intersection}).

\begin{equation}
    \label{eq:ray_plane_intersection}
    \Tilde{c}^{k\times k} = \frac{n4}{n1 \cdot u_{i} + n2 \cdot v_{i} + n3}
\end{equation}

where ($n_{1}, n_{2}, n_{3}, n_{4}$) describe the previously estimated plane coefficients and $u_{i}, v_{i}$ denote $k\times k$ patch-wise normalized coordinates of pixel $i$.

\subsubsection{Attention mechanism}
Convolutional neural networks have shown outstanding performance in enhancing local feature representations. However, focusing on relevant features while suppressing irrelevant ones further improves the process of capturing the visual structure of a scene. The convolutional block attention module (CBAM) proposed in~\citep{woo2018} sequential applies a channel attention module and a spatial attention module to a given feature map. The process consists of two sequential element-wise multiplications. The first one multiplies the input feature map ($\mathcal{F}$) and the channel attention map ($M_{c}$). The second one is performed between the output of the first multiplication ($\mathcal{F}'$) and the output of the spatial attention module ($M_{s}$), resulting in a refined feature map $\mathcal{F}''$.
The whole process can be summarised as shown in Eq.(~\ref{eq:CBAM}): 

\begin{empheq}[left=\empheqlbrace]{align}
  \label{eq:CBAM}
    & \mathcal{F}' = \mathcal{F} \otimes M_{c}(\mathcal{F}) \\
    & \mathcal{F}'' = \mathcal{F}' \otimes M_{s}(\mathcal{F}') \nonumber 
\end{empheq}
    
\begin{figure*}[!t]
    \centering
    \includegraphics[width=\linewidth]{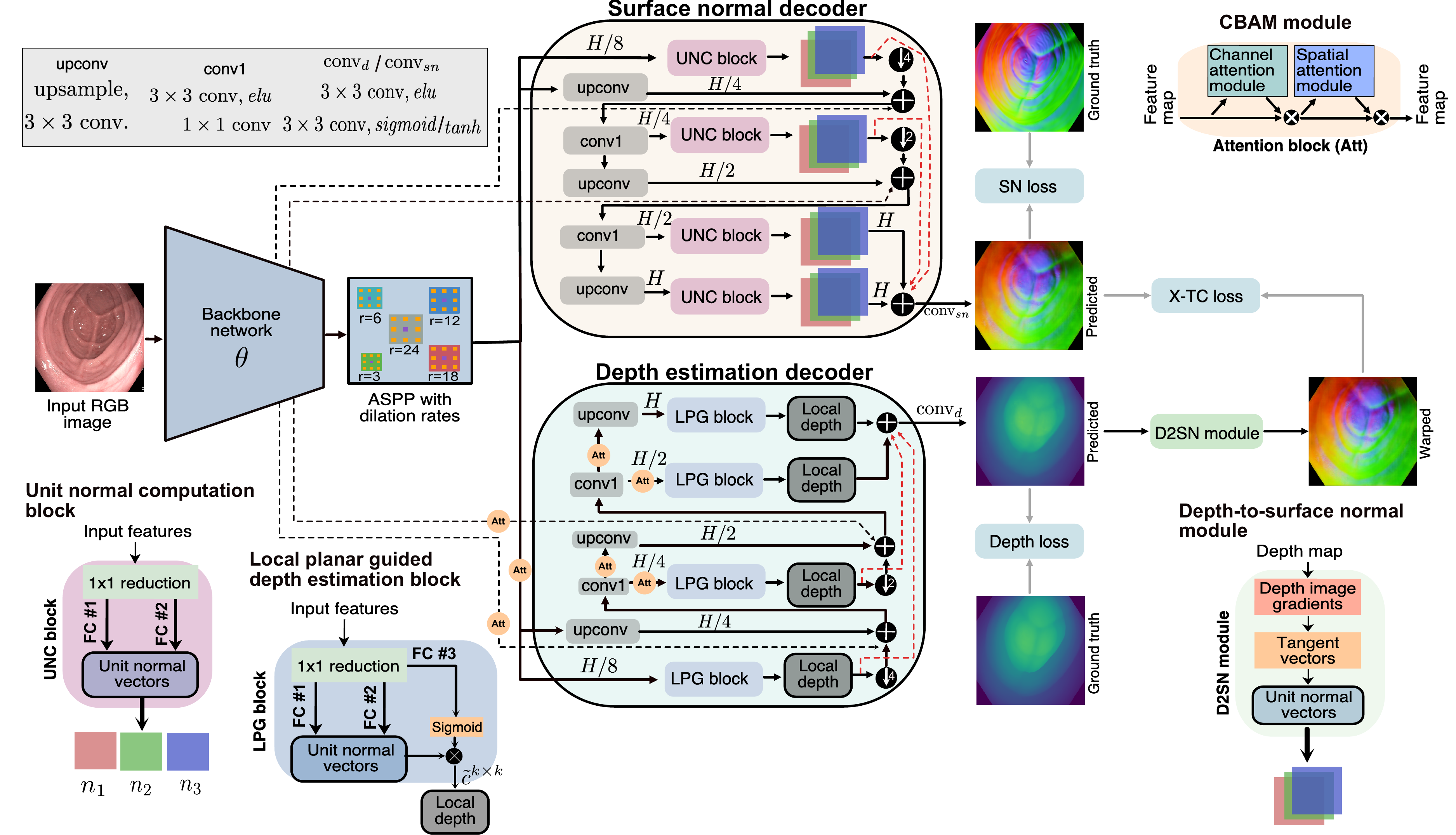}
    \caption{Multi-task learning with cross-task consistency (Col3D-MTL). Our proposed framework follows the encoder-decoder scheme, in which the encoder consists of a shared backbone, $\theta$, followed by an atrous spatial pyramid (ASPP) module to extract contextual information at different dilation rates. The decoder stage comprises a primary depth estimator decoder (bottom) and an auxiliary surface normal decoder (top). Our unit normal computation block (UNC block) uses two feature channels (FC) to compute unit normal vectors ($n_{1}$, $n_{2}$, and $n_{3}$). The local planar guided depth estimation block (LPG block) also uses a third FC to compute the perpendicular distance to the camera, which is incorporated together with the unit normals to provide local depth information $\Tilde{c}^{k\times k}$ by the ray-plane intersection (see Eq.(~\ref{eq:ray_plane_intersection})). CBAM modules (Att) are introduced at the skip connections and after the convolutional layers of the depth decoder to enhance global context awareness. The depth-to-surface normal (D2SN) module receives the predicted depth and outputs a warped surface normal map, which is compared against the surface normal prediction to enforce consistency among tasks.
    The surface normal decoder and D2SN module can be easily combined with other depth estimators for an end-to-end MTL-X-TC.}
    \label{fig:multitask_framework}
\end{figure*}

This module has been used by~\cite{li2023} in the skip connections and in the decoder of their network to recover meaningful global information at a low computational cost. We followed a similar approach to~\citep{li2023} by incorporating CBAM modules in the skip connections and at each resolution level of our depth decoder. We aim to leverage the local feature representation of convolutional neural networks to extract monocular depth cues and the global context awareness of CBAM modules to relate them effectively. The incorporation of CBAM modules into our framework is shown in Fig.~\ref{fig:multitask_framework}.

\subsubsection{Multi-task learning network with UNC block}
Our proposed framework is based on the geometric relationship between the depth and surface normal information of a 3D scene. Following this statement, our method aims to improve its depth estimation robustness by incorporating a geometrically related auxiliary task. The proposed architecture consists of a single shared encoder and two independent decoders. The purpose of the shared encoder is to extract meaningful geometric features that represent the 3D scene, while the two decoders are used to regress depth and surface normal maps.

Our framework extends the depth estimation network by adding a surface normal decoder located at the top of Fig.~\ref{fig:multitask_framework}. Within our surface normal decoder, we introduce the unit normal computation block (UNC block) to compute surface normal maps at multiple resolutions. Our UNC block is based on the LPG blocks without the computation of the ray-plane intersection. Each surface normal map is represented by an RGB image, where each channel represents one particular axis: the red channel denotes the $x$-axis, the green channel represents the $y$-axis, and the blue channel represents the $z$-axis. The outputs of all UNC blocks undergo a channel concatenation followed by convolutional layers to compute the final surface normal prediction.

The multi-task learning framework combines the losses of both tasks as described in Eq.(~\ref{eq:multitask_loss}).
\begin{equation}
    \label{eq:multitask_loss}
    \mathcal{L}_{MTL} = \lambda_{1} \cdot \mathcal{L}_{depth} + \lambda_{2} \cdot \mathcal{L}_{sn}
\end{equation}
where $\lambda_{1}, \lambda_{2}$ are weighting factors equally set to 0.5, $\mathcal{L}_{depth}$ represents the scale-invariant logarithmic error (SILog) loss between the predicted and ground truth depth maps, and $\mathcal{L}_{sn}$ symbolises the mean absolute error (MAE) loss between the estimated surface normal and its corresponding ground truth.

\subsubsection{Cross-task consistency loss with D2SN module}
We incorporate a cross-task consistency (X-TC) loss into our multi-task learning framework to enforce consistency among depth and surface normal predictions. To this end, we add a depth-to-surface normal (D2SN) warping module based on the mathematical method proposed by~\cite{nakagawa2015}. Our warping module uses the output of our depth estimation decoder. The predicted depth map is processed within this module to generate a warped surface normal representation. The output of this module introduces a consistency constraint by being compared against the prediction of the surface normal decoder.

\paragraph{Depth-to-surface normal module (D2SN module)}
Since normal estimation is equivalent to fitting a plane to a local point cloud in the 3D space, several approaches leveraging optimisation techniques have been proposed. However, these methods are expensive in terms of computational resources. Therefore, our D2SN module, which is shown in the right section of Fig.~\ref{fig:multitask_framework}, computes a surface normal from depth image gradients (DIG) as shown in~\citep{nakagawa2015}. The authors consider that adjacent 3D points in a depth image can be used to compute a local 3D plane whose orthogonal vector is equivalent to the normal vector at that particular pixel. Their proposed method consists of three steps:

\begin{enumerate}
    \item Depth image gradients: Given the location of a pixel ($x,y$) and its depth value ($Z$), pixels can be projected to the 3D space P$(X, Y, Z)$ by a transformation matrix with known camera intrinsic parameters. The generated point cloud is used to compute the partial directional derivatives as shown in Eq.~(\ref{eq:partial_derivatives}):
        \begin{empheq}[left=\empheqlbrace]{align}
        \label{eq:partial_derivatives_z}
            &\frac{\partial{Z(x,y)}}{\partial{x}} = Z(x+1,y) - Z(x,y) \nonumber \\
            &\frac{\partial{Z(x,y)}}{\partial{y}} = Z(x,y+1) - Z(x,y) \nonumber
        \end{empheq}
        \begin{empheq}[left=\empheqlbrace]{align}
        \label{eq:partial_derivatives}
            &\frac{\partial{X(x,y)}}{\partial{x}} = \frac{Z(x,y)}{f} + \frac{(x-c_{x})}{f}\frac{\partial{Z(x,y)}}{\partial{x}} \\
            &\frac{\partial{X(x,y)}}{\partial{y}} = \frac{(x-c_{x})}{f}\frac{\partial{Z(x,y)}}{\partial{y}} \nonumber \\
            &\frac{\partial{Y(x,y)}}{\partial{x}} = \frac{(x-c_{x})}{f}\frac{\partial{Z(x,y)}}{\partial{y}} \nonumber \\
            &\frac{\partial{Y(x,y)}}{\partial{y}} = \frac{Z(x,y)}{f} + \frac{(y-c_{y})}{f}\frac{\partial{Z(x,y)}}{\partial{y}} \nonumber 
        \end{empheq}
    \item Tangent vectors: The $x$ and $y$ directional derivatives at a given 3D point, P, can be used as tangent vectors of the surface as shown in Eq.(~\ref{eq:tangent_vectors}):
        \begin{empheq}[left=\empheqlbrace]{align}
        \label{eq:tangent_vectors}
            & \mathbf{v}_{x}(x,y) = \frac{\partial{X}(x,y)}{\partial{x}}, \frac{\partial{Y}(x,y)}{\partial{x}}, \frac{\partial{Z(x,y)}}{\partial{x}} \\
            & \mathbf{v}_{y}(x,y) = \frac{\partial{X}(x,y)}{\partial{y}}, \frac{\partial{Y}(x,y)}{\partial{y}}, \frac{\partial{Z(x,y)}}{\partial{y}} \nonumber
        \end{empheq}
    \item Normal vector: The cross-product of the two tangent vectors described in the previous step is calculated to get the normal vector as in Eq.(~\ref{eq:cross_product}):
        \begin{equation}
        \label{eq:cross_product}
            \mathbf{n}(x,y) = \mathbf{v}_{x}(x,y) \times \mathbf{v}_{y}(x,y)
        \end{equation}
\end{enumerate}

Our multi-task learning with cross-task consistency framework combines the losses from both tasks and the cross-task consistency loss into the final weighted loss function described in Eq.(~\ref{eq:multitask_XTC_loss}).
\begin{equation}
    \label{eq:multitask_XTC_loss}
    \mathcal{L}_{final} = \lambda_{1} \cdot \mathcal{L}_{depth} + \lambda_{2} \cdot \mathcal{L}_{sn} + \lambda_{3} \cdot \mathcal{L}_{X-TC}
\end{equation}
where $\lambda_{1}, \lambda_{2}$, and $\lambda_{3}$ are weighting factors, and $\mathcal{L}_{X-TC}$ represents the root mean squared error (RMSE) loss between the predicted and warped surface normal maps. Eq.(~\ref{eq:losses}) defines each of the loss functions used in our final loss.

\begin{empheq}[left=\empheqlbrace]{align}
  \label{eq:losses}
    & L_{depth} = \alpha \sqrt{D(g)} ;\: D(g) = \frac{1}{N}\sum_{i}g_{i}^{2} - \frac{\lambda}{N}^{2}\left(\sum_{i}g_{i}\right)^{2}  \\
    & g_{i} = log \left(y_{i\_depth}\right)  - log \left(\hat{y}_{i\_depth}\right) \nonumber \\
    & L_{sn} = \frac{1}{N}\sum_{i}|y_{i\_sn} - \hat{y}_{i\_sn}| \nonumber \\
    & L_{x-tc} = \sqrt{\frac{1}{N}\sum_{i}(y_{i\_sn}^{*} - \hat{y}_{i\_sn})^{2}} \nonumber
\end{empheq}
where $\lambda=0.85$, $y_{i}$ represents the ground truth map, $\hat{y_{i}}$ denotes the prediction map, and $y^{*}_{i\_sn}$ the warped surface normal map.

\section{Experiments and Results}{\label{sec:resultsandExperiments}}

\subsection{Training setup}
Each model presented in this study is trained using a single NVIDIA V100 GPU. All models are trained for up to 50 epochs (pix2pix and MonoDepth+FPN are trained for 200 epochs) using a batch size of 8, with an initial learning rate of $1e^{-4}$, and a weight decay of $1e^{-2}$. The input images are resized from their original resolution to $320 \times 320$ pixels.
Only random rotation is performed as a data augmentation technique to avoid the loss of structural information and visual cues. Random cropping, which is suggested by baseline methods, is discarded because we observed that it could lead to heavy close-ups towards the walls of the colon, drastically reducing contextual information.

\subsection{Ablation study setup}
We perform an ablation study to analyse the proposed network components that lead to the design of our multi-task learning approach with cross-task consistency featuring attention mechanisms. We select the BTS architecture proposed by~\cite{lee2019} as our baseline method. Before incorporating our multi-task learning approach, we add CBAM attention modules at different stages of the depth decoder and its skip connections. We further extend this method following a multi-task learning scheme. Experimentally, we define the best loss function to optimise our auxiliary task. Additionally, we implement a cross-task consistency module into our multi-task learning approach. Finally, we conduct a hyperparameter study on our multi-task learning with a cross-task consistency framework to determine the set of $\lambda$ values in our loss function that leads to the best trade-off performance among both tasks.
In Table~\ref{tab:ablation_setup}, we describe the different network configurations trained and evaluated in our ablation study.

\begin{table}[t!]
\caption{Model configurations. Our ablation study setup is constituted by four different model configurations.}
\label{tab:ablation_setup}
\centering
\resizebox{\textwidth}{!}{
\begin{tabular}{llll}
\toprule
\textbf{\scriptsize Model ID} & \textbf{\scriptsize CBAM} & \textbf{\scriptsize MTL} & \textbf{\scriptsize X-TC} \\ \hline \hline
\scriptsize BTS (baseline)~\citep{lee2019} & & & \\
\scriptsize BTS-CBAM              & \scriptsize \ding{51} & & \\
\scriptsize BTS-CBAM-MTL        & \scriptsize \ding{51} & \scriptsize \ding{51} & \\
\scriptsize BTS-CBAM-MTL-X-TC (Col3D-MTL) & \scriptsize \ding{51} & \scriptsize \ding{51} & \scriptsize \ding{51} \\ 
\bottomrule
\end{tabular}}
\end{table}

\subsection{Metrics and assessment}
To evaluate our methods, we follow standard depth estimation metrics described by~\cite{eigen2014}. These include five error metrics: absolute relative error (Abs Rel),  squared relative error (Sq Rel), logarithmic error (log10), root mean squared error (RMSE), root mean squared logarithmic error (RMSE$_{log}$), scale-invariant logarithmic error (SILog); and three accuracy metrics: $\delta_{1}$, $\delta_{2}$, and $\delta_{3}$.
We used standard surface normal evaluation metrics, which include two error metrics: mean angular error (Mean Ang) and median angular error (Median Ang), and three accuracy metrics: $\delta_{11.25^{\circ}}$, $\delta_{22.5^{\circ}}$, and $\delta_{30^{\circ}}$

\subsection{Results}
In this section, we provide the quantitative and qualitative results from our network configuration ablation study, followed by our hyperparameter study on the $\lambda$ weighting factors of our final loss function. Finally, we compare our approach against other state-of-the-art methods.

\subsubsection{Quantitative results}

In Table~\ref{tab:MTL-results}, we provide the results of our ablation study on the proposed network configurations. We compare the performances of BTS, BTS-CBAM, BTS-CBAM-MTL, and BTS-CBAM-MTL-X-TC for all depth and surface normal evaluation metrics on the validation set of the C3VD dataset. Our BTS-CBAM network yields a relative improvement over the BTS method by 8.9\% and 2.4\% in terms of SILog and $\delta_{1}$, respectively.
\begin{table*}[th!]
\centering
\caption{Quantitative results for various network configurations on validation set (ablation study). Each network uses the learned weights that lead to a better performance during the training stage. \textbf{First} and \underline{second} best performing methods for each evaluation metric are formatted.}
\label{tab:MTL-results}
\resizebox{\textwidth}{!}{
\begin{tabular}{ll|ccccccccc|ccccc}
\toprule
\multicolumn{1}{l}{\textbf{Method}} & \multicolumn{1}{c|}{\textbf{Losses}} & \textbf{Abs Rel $\downarrow$} & \textbf{Sq Rel $\downarrow$} & \textbf{log 10 $\downarrow$} & \textbf{RMSE $\downarrow$} & \textbf{RMSE$_{log}$ $\downarrow$} & \textbf{SILog $\downarrow$} & \textbf{$\delta_{1}$ $\uparrow$} & \textbf{$\delta_{2}$ $\uparrow$} & \textbf{$\delta_{3}$ $\uparrow$} & \multicolumn{1}{l}{\textbf{Mean Ang $\downarrow$}} & \multicolumn{1}{l}{\textbf{Median Ang $\downarrow$}} & \multicolumn{1}{l}{$\delta_{11.25^{\circ}}$ $\uparrow$} & \multicolumn{1}{l}{\textbf{$\delta_{22.5^{\circ}}$ $\uparrow$}} & \multicolumn{1}{l}{\textbf{$\delta_{30^{\circ}}$ $\uparrow$}} \\ \hline \hline
BTS (baseline) & Depth: SILog & 0.179 & 1.088 & 0.073 & 5.667 & 0.208 & 14.066 & 0.756 & 0.959 & 0.992 & - & - & - & - & - \\ 
BTS-CBAM     & Depth: SILog & 0.170 & 0.979 & 0.070 & 5.503 & 0.198 & \underline{12.814} & 0.774 & 0.969 & 0.994 & - & - & - & - & - \\ \hline
\multirow{3}{*}{\begin{tabular}[c]{@{}l@{}}BTS-CBAM-\\MTL\end{tabular}} & \begin{tabular}[c]{@{}l@{}}Depth: SILog\\ SN: $L_{1}$\end{tabular} & 0.166 & 0.934 & 0.070 & 5.422 & 0.195 & 12.872 & 0.785 & 0.969 & \underline{0.996} & 43.892 & 35.363 & \underline{17.637} & \underline{40.340} & 51.814 \\ \cline{2-16} 
                 & \begin{tabular}[c]{@{}l@{}}Depth: SILog\\ SN: $L_{2}$\end{tabular} & \underline{0.163} & \underline{0.900} & \underline{0.067} & \underline{5.298} & \underline{0.190} & 13.219 & \underline{0.797} & \underline{0.972} & 0.995 & 44.337 & 35.142 & 13.940 & 37.288 & 49.594   \\ \hline
\multirow{3}{*}{\begin{tabular}[c]{@{}l@{}}BTS-CBAM-\\MTL-X-TC\\(Col3D-MTL) \end{tabular}}
& \begin{tabular}[c]{@{}l@{}l@{}}Depth: SILog\\ SN: $L_{1}$\\ X-TC: $L_{2}$\end{tabular} & \textbf{0.156} & \textbf{0.805} & \textbf{0.065} & \textbf{4.994} & \textbf{0.186} & \textbf{12.530} & \textbf{0.809} & \textbf{0.975} & \textbf{0.997} & \textbf{23.999} & \textbf{21.382} & \textbf{24.232} & \textbf{55.883} & \textbf{71.758} \\ \cline{2-16} 
              & \begin{tabular}[c]{@{}l@{}l@{}}Depth: SILog\\ SN: $L_{2}$\\ X-TC: $L_{2}$\end{tabular} & 0.176 & 1.057 & 0.073 & 5.723 & 0.211 & 14.822 & 0.731 & 0.968 & \underline{0.996} & \underline{40.664} & \underline{32.630} & 15.974 & 40.219 & \underline{52.394}   \\
\bottomrule
\end{tabular}
}
\end{table*}
\begin{table*}[t!]
\centering
\caption{Hyperparameter study on validation set. {Proposed BTS-CBAM-MTL-X-TC} networks are evaluated in the validation set to avoid $\lambda$ weighting factors to be adjusted based on the testing data. Each network uses the learned weights that lead to a better performance during the training stage. \textbf{First} and \underline{second} best performing methods for each evaluation metric are formatted.}
\label{tab:MTL-XTC-results}
\resizebox{\textwidth}{!}{
\begin{tabular}{lccc|ccccccccc|ccccc}
\toprule
\multicolumn{1}{l|}{\textbf{Method}}                     & \textbf{$\lambda_{1}$} & \multicolumn{1}{l}{$\lambda_{2}$} & \multicolumn{1}{l|}{\textbf{$\lambda_{3}$}} & \textbf{Abs Rel $\downarrow$} & \textbf{Sq Rel $\downarrow$} & \textbf{log 10 $\downarrow$} & \textbf{RMSE $\downarrow$} & \textbf{RMSE$_{log}$ $\downarrow$} & \textbf{SILog $\downarrow$} & \textbf{$\delta_{1}$ $\uparrow$} & \textbf{$\delta_{2}$ $\uparrow$} & \textbf{$\delta_{3}$ $\uparrow$} & \multicolumn{1}{l}{\textbf{Mean Ang $\downarrow$}} & \multicolumn{1}{l}{\textbf{Median Ang $\downarrow$}} & \multicolumn{1}{l}{$\delta_{11.25^{\circ}}$ $\uparrow$} & \multicolumn{1}{l}{\textbf{$\delta_{22.5^{\circ}}$ $\uparrow$}} & \multicolumn{1}{l}{\textbf{$\delta_{30^{\circ}}$ $\uparrow$}} \\ \hline \hline
\multicolumn{1}{l|}{\multirow{8}{*}{\begin{tabular}[c]{@{}l@{}}BTS-CBAM-\\MTL-X-TC\\(Col3D-MTL)\end{tabular}}} & 0.4 & 0.3 & 0.3 & 0.178 & 1.061 & 0.072 & 5.639 & 0.204 & 13.632 & 0.769 & 0.963 & 0.994 & \textbf{22.957} & \textbf{20.405} & \textbf{25.028} & \textbf{57.662} & \textbf{73.668} \\
\multicolumn{1}{l|}{}                                   & 0.4 & 0.2 & 0.2 & 0.187 & 1.116 & 0.077 & 5.664 & 0.213 & 13.672 & 0.726 & 0.965 & 0.990 & 30.100 & 24.999 & 21.081 & 49.877 & 63.892 \\
\multicolumn{1}{l|}{}                                   & 0.5 & 0.3 & 0.2 & \textbf{0.156} & \textbf{0.805} & \textbf{0.065} & \underline{4.994} & \textbf{0.186} & \textbf{12.530} & \textbf{0.809} & \textbf{0.975} & \textbf{0.997} & 23.999 & \underline{21.382} & \underline{24.232} & \underline{55.883} & 71.758 \\
\multicolumn{1}{l|}{}                                   & 0.5 & 0.4 & 0.1 & \underline{0.161} & \underline{0.837} & \underline{0.066} & \textbf{4.901} & \underline{0.189} & \underline{13.036} & \underline{0.795} & \underline{0.974} & \textbf{0.997} & 34.013 & 24.606 & 21.812 & 50.571 & 63.335 \\
\multicolumn{1}{l|}{}                                   & 0.6 & 0.2 & 0.2 & 0.222 & 1.470 & 0.089 & 6.269 & 0.248 & 16.163 & 0.620 & 0.939 & 0.987 & 29.338 & 25.449 & 17.860 & 45.908 & 62.025 \\
\multicolumn{1}{l|}{}                                   & 0.6 & 0.3 & 0.1 & 0.187 & 1.083 & 0.075 & 5.598 & 0.209 & 13.773 & 0.742 & 0.957 & 0.994 & 45.448 & 35.227 & 15.080 & 36.828 & 48.007 \\
\multicolumn{1}{l|}{}                                   & 0.7 & 0.2 & 0.1 & 0.178 & 1.028 & 0.070 & 5.128 & 0.202 & 14.329 & 0.764 & 0.961 & 0.991 & 33.879 & 28.175 & 17.835 & 43.982 & 58.190 \\
\multicolumn{1}{l|}{}                                   & 0.8 & 0.1 & 0.1 & 0.180 & 1.078 & 0.074 & 5.753 & 0.210 & 13.573 & 0.739 & 0.963 & \underline{0.995} & \underline{23.880} & 21.434 & 23.147 & 55.364 & \underline{71.800} \\
\bottomrule
\end{tabular}
}
\end{table*}
Following the incorporation of attention mechanisms, the middle section of Table~\ref{tab:MTL-results} illustrates the effect of the $L_{1}$ and $L_{2}$ loss functions to optimise the surface normal decoder of our BTS-CBAM-MTL approach. The use of the $L_{1}$ loss function leads to a relative improvement of 2.6\% on SILog metric but a relative decrease of 1.5\% on $\delta_{1}$ regarding the use of the $L_{2}$ loss. Furthermore, our surface normal decoder optimised through the $L_{1}$ loss demonstrates a relative improvement of 1\% and 26.5\% on mean angular error and $\delta_{11.25^{\circ}}$, respectively. 
The bottom section of Table~\ref{tab:MTL-results} assesses the performance of our BTS-CBAM-MTL-X-TC employing the $L_{1}$ and $L_{2}$ loss functions to optimise our surface normal decoder. Initially, the $\lambda$ weighting factors in our final loss function (Eq.(~\ref{eq:multitask_XTC_loss})) are set to $\lambda_{1}=0.5$, $\lambda_{2}=0.3$, and $\lambda_{3}=0.2$. Our proposed network configuration optimising our surface normal decoder with the $L_{1}$ loss outperforms all previous network configurations in all depth and surface normal evaluation metrics.

Table~\ref{tab:MTL-XTC-results} includes the results of our hyperparameter study on different sets of $\lambda$ weighting factors in the loss function (Eq.(~\ref{eq:multitask_XTC_loss})) of our BTS-CBAM-MTL-X-TC network. Based on our experimental results, the best $\lambda$ configuration set consists of $\lambda_{1} = 0.5$, $\lambda_{2} = 0.3$, and $\lambda_{3} = 0.2$. This configuration leads to the best performance among all evaluated models, achieving a relative improvement of 10.9\%, 11.9\%, and 7\% on SILog, RMSE, and $\delta_{1}$ metrics over our baseline method. Furthermore, we outperform the surface normal prediction of our BTS-CBAM-MTL configuration by a relative improvement of 45.3\%, 39.5\%, and 37.4\% on mean angular error, median angular error, and $\delta_{11.25^{\circ}}$.

In Table~\ref{tab:comparative-results}, we compare the performance of our baseline and our Col3D-MTL methods against other state-of-the-art methods on supervised monocular depth estimation: NeWCRFS~\citep{chen2020}, pix2pix~\citep{isola2017,rau2019}, and MonoDepth+FPN~\citep{ali2021}. MonoDepth+FPN~\citep{ali2021} is used to estimate depth maps on oesophageal endoscopy, while pix2pix~\citep{rau2019} is applied on a synthetic colonoscopy dataset. All methods are trained on the same data distribution and tested on the same held-out testing data, setting a new benchmark on the C3VD dataset. Our proposed framework achieves the best performance on all depth evaluation metrics. It improves our baseline method by 3.2\%, 12.6\%, and 10.4\% on SILog, RMSE and $\delta_{1}$, respectively. Moreover, it outperforms the state-of-the-art method NeWCRFs by  6.6\%, 0.2\%, and 4.9\% on the same evaluation metrics.

\begin{table*}[t!]
\centering
\caption{Benchmark results on test set. Evaluation of three state-of-the-art methods, our baseline method and our proposed framework on the C3VD dataset. All methods are trained and evaluated on the same data distributions. \textbf{First} and \underline{second} best performing methods for each evaluation metric are formatted.}
\label{tab:comparative-results}
\resizebox{\textwidth}{!}{
\begin{tabular}{lcllccccccccc}
\toprule
\multicolumn{1}{l|}{\textbf{Method}}    &  \textbf{Abs Rel $\downarrow$} & \textbf{Sq Rel $\downarrow$} & \textbf{log 10 $\downarrow$} & \textbf{RMSE $\downarrow$} & \textbf{RMSE}$_{log}$ $\downarrow$ & \textbf{SILog $\downarrow$} & \textbf{$\delta_{1}$ $\uparrow$} & \textbf{$\delta_{2}$ $\uparrow$} & \textbf{$\delta_{3}$ $\uparrow$}  \\ \hline \hline
\multicolumn{1}{l|}{\begin{tabular}[c]{@{}l@{}}pix2pix\\~\citep{isola2017,rau2019} \end{tabular}} & 0.157$\pm$0.103 & 0.730$\pm$0.647 & 0.062$\pm$0.034 & 3.721$\pm$1.385 & 0.237$\pm$0.144 & 21.062$\pm$14.118 & 0.801$\pm$0.181 & 0.956$\pm$0.078 & 0.986$\pm$0.029 \\
\multicolumn{1}{l|}{\begin{tabular}[c]{@{}l@{}}MonoDepth+FPN\\~\citep{ali2021} \end{tabular}}       & 0.204$\pm$0.052 & 1.494$\pm$0.832 & 0.090$\pm$0.025 & 6.762$\pm$1.812 & 0.342$\pm$0.078 & 31.737$\pm$6.25 & 0.628$\pm$0.176 & 0.949$\pm$0.046 & 0.991$\pm$0.010  \\
\multicolumn{1}{l|}{\begin{tabular}[c]{@{}l@{}}NeWCRFs\\~\citep{chen2020} \end{tabular}}               & 0.133$\pm$0.097 & \underline{0.557$\pm$0.595} & \underline{0.048$\pm$0.033} & \underline{3.058$\pm$1.419} & 0.151$\pm$0.082 & 11.817$\pm$4.714 & \underline{0.854$\pm$0.197} & 0.973$\pm$0.066 & 0.997$\pm$0.006  \\
\multicolumn{1}{l|}{\begin{tabular}[c]{@{}l@{}}BTS\\~\citep{lee2019} \end{tabular}}                   & \underline{0.127$\pm$0.089} & 0.622$\pm$0.543 & 0.055$\pm$0.034 & 3.823$\pm$2.024 & \underline{0.150$\pm$0.089} & \underline{11.400$\pm$5.614} & 0.812$\pm$0.198 & \underline{0.979$\pm$0.051} & \underline{0.998$\pm$0.003}  \\
\multicolumn{1}{l|}{\begin{tabular}[c]{@{}l@{}}\textbf{BTS-CBAM-MTL-X-TC}\\\textbf{(Col3D-MTL, ours)} \end{tabular}}                  & \textbf{0.109$\pm$0.064} & \textbf{0.386$\pm$0.316} & \textbf{0.046$\pm$0.024} & \textbf{3.052$\pm$1.143} & \textbf{0.131$\pm$0.065} & \textbf{11.035$\pm$4.856} & \textbf{0.896$\pm$0.140} & \textbf{0.989$\pm$0.021} & \textbf{0.998$\pm$0.003}  \\
\bottomrule
\end{tabular}
}
\end{table*}

Table~\ref{tab:per-segment} analyses the performance of all the evaluated methods on each colon segment of the testing set separately. 
On all colon segments, except for the caecum, Col3D-MTL scores higher than all the other methods in terms of $\delta_{1}$. 
On the caecum and transverse segments, which respectively represent 43\% and 36\% of our training data, our baseline and our proposed framework achieve the first and second best-performing methods in terms of SILog, RMSE, and $\delta_{1}$. Only NeWCRFs accomplishes a lower RMSE value on the caecum segment, achieving a relative improvement of 4.9\% over our baseline.
Considering the sigmoid segment, which only constitutes 21\% of the training set, our approach outperforms our baseline method by a relative improvement of 5\%, 31.1\%, and 22.7\% on SILog, RMSE, and $\delta_{1}$. However, our network is only surpassed by NeWCRFs on SILog and RMSE metrics.
The most remarkable improvement occurs in the descending segment of the colon, which was not given during the training stage. On this colon segment, Col3D-MTL obtains the best-performing metrics among all the evaluated methods while achieving a relative improvement of 9.5\%, 16.4\%, and 32.3\% on SILog, RMSE, and $\delta_{1}$ over our baseline.

\begin{table}[!th]
\centering
\caption{Quantitative results for per colon segment. Evaluation of all evaluated methods on each colon segment in the testing set is provided. \textbf{First} and \underline{second} best performing method for each evaluation metric on each colon segment is formatted. In '(.)' we include the percentage of training samples used from each segment.}
\label{tab:per-segment}
\resizebox{\textwidth}{!}{
\begin{tabular}{ll|ccc}
\toprule
\textbf{Colon segment}            & \textbf{Method} & \textbf{SILog $\downarrow$} & \textbf{RMSE $\downarrow$} & \textbf{$\delta_{1} \uparrow$} \\ \hline \hline
\multirow{5}{*}{\begin{tabular}[c]{@{}l@{}}Caecum\\(43\%)\end{tabular}} 
& pix2pix & 6.805$\pm$1.345 & 2.649$\pm$1.342 & 0.956$\pm$0.083 \\ 
& MonoDepth+FPN & 28.834$\pm$4.958 & 7.143$\pm$1.405 & 0.822$\pm$0.052 \\ 
& NeWCRFs & 7.494$\pm$0.696 & \textbf{1.567$\pm$0.552} & \underline{0.996$\pm$0.004} \\ 
& BTS & \textbf{5.171$\pm$1.065} & \underline{1.649$\pm$0.447} & \textbf{0.997$\pm$0.003} \\
& Ours & \underline{5.738$\pm$1.232} & 2.069$\pm$0.547 & \underline{0.996$\pm$0.004} \\ \hline
\multirow{5}{*}{\begin{tabular}[c]{@{}l@{}}Transverse\\(36\%)\end{tabular}}
& pix2pix & 13.464$\pm$1.304 & 2.374$\pm$0.100 & 0.929$\pm$0.017 \\ 
& MonoDepth+FPN & 25.409$\pm$0.695 & 4.894$\pm$0.279 & 0.759$\pm$0.026 \\ 
& NeWCRFs & 8.858$\pm$0.917 & 2.598$\pm$0.416 & 0.992$\pm$0.004 \\ 
& BTS & \textbf{6.243$\pm$0.265} & \underline{1.840$\pm$0.145} & \underline{0.997$\pm$0.002} \\
& Ours & \underline{6.412$\pm$0.490} & \textbf{1.648$\pm$0.056} & \textbf{0.997$\pm$0.001} \\ \hline
\multirow{5}{*}{\begin{tabular}[c]{@{}l@{}}Sigmoid\\(21\%)\end{tabular}}
& pix2pix & 32.429$\pm$12.481 & 4.618$\pm$0.593 & 0.745$\pm$0.068 \\ 
& MonoDepth+FPN & 36.365$\pm$4.780 & 6.422$\pm$0.546 & 0.459$\pm$0.072 \\ 
& NeWCRFs & \textbf{13.135$\pm$1.600} & \textbf{3.664$\pm$1.052} & \underline{0.809$\pm$0.158} \\ 
& BTS & 16.276$\pm$1.526 & 5.838$\pm$0.498 & 0.715$\pm$0.048  \\
& Ours & \underline{15.460$\pm$1.240} & \underline{4.020$\pm$0.353} & \textbf{0.877$\pm$0.023} \\ \hline
\multirow{5}{*}{\begin{tabular}[c]{@{}l@{}}Descending\\(0\%)\end{tabular}} 
& pix2pix & 26.52$\pm$6.868 & \underline{5.095$\pm$0.599} & 0.432$\pm$0.121 \\ 
& MonoDepth+FPN & 31.629$\pm$6.907 & 11.405$\pm$1.479 & \underline{0.523$\pm$0.109} \\ 
& NeWCRFs & 22.137$\pm$2.915 & 5.468$\pm$0.437 & 0.464$\pm$0.098 \\ 
& BTS & \underline{15.647$\pm$2.705} & 5.121$\pm$0.627 & 0.396$\pm$0.138  \\
& Ours & \textbf{14.156$\pm$2.325} & \textbf{4.279$\pm$0.584} & \textbf{0.524$\pm$0.144} \\
\bottomrule
\end{tabular}}
\end{table}

\subsubsection{Qualitative results}
\begin{figure*}[!th]
    \centering
    \includegraphics[width=\linewidth]{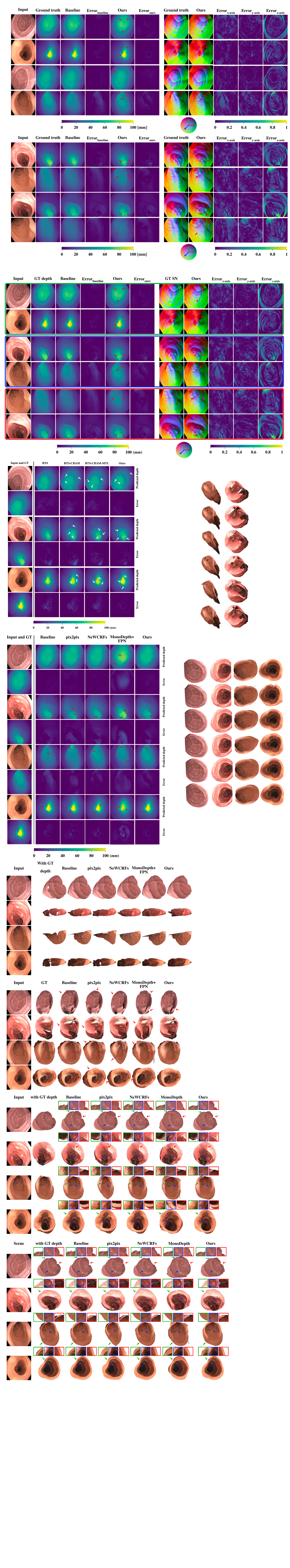}
    \caption{Qualitative comparison between our baseline and our proposed framework on best, average, and worst performing cases. We show absolute error maps for both methods to observe the most challenging regions and the impact of our BTS-CBAM-MTL-X-TC. In the first two rows, we can observe the best-performing cases (green) in which both methods lead to low absolute error maps. The third and fourth rows show average-performing cases (blue) in which the lack of texture and regions with high-depth variability affecting our baseline method are addressed by our BTS-CBAM-MTL-X-TC framework. Consistency can be observed on the predicted surface normal maps, which help to recover the shape of the scene, e.g., folds of the colon and small protuberances like polyps. The last two rows represent challenging cases (red) in which both methods generate less accurate depth estimations. Low lighting regions, usually located at the furthest section of the scene, represent challenging cases for both methods.}
    \label{fig:qualitative_cases}
\end{figure*}

Fig.~\ref{fig:qualitative_cases} contains sample input images with their corresponding ground truth annotations and the predictions of our baseline and our proposed method.
We can observe that our baseline properly recovers a global depth map of the 3D scene; however, it has a tendency to generate smooth transitions between anatomical structures. Our proposed method addresses these cases by recovering geometrical information about the scene and enforcing cross-task consistency between our depth and surface normal predictions, leading to sharper boundaries and reduced visual artefacts, e.g., specular reflection.
We include an absolute error map for each channel pair among prediction and ground truth maps from both tasks. From the absolute error maps, we can observe that our baseline generates brighter regions than Col3D-MTL, i.e., our approach recovers the 3D information of the scene with less absolute error than its baseline method.

By analysing the most challenging samples, we can observe that the cases in which our methods fail to recover an accurate depth estimation are small regions with low lighting conditions. Even though our surface normal decoder recovers the overall geometry of small structures, such as polyps, it does not compute a detailed representation of the surface orientation within these small regions. 
Looking at the areas denoted with red arrows, we can observe that regions with higher errors in our depth maps are consistent with regions with higher errors in our surface normal maps. Despite these focalised errors, the overall depth estimation of our proposed method performs better than the one from its baseline.

\begin{figure}[!th]
    \centering
    \includegraphics[width=\linewidth]{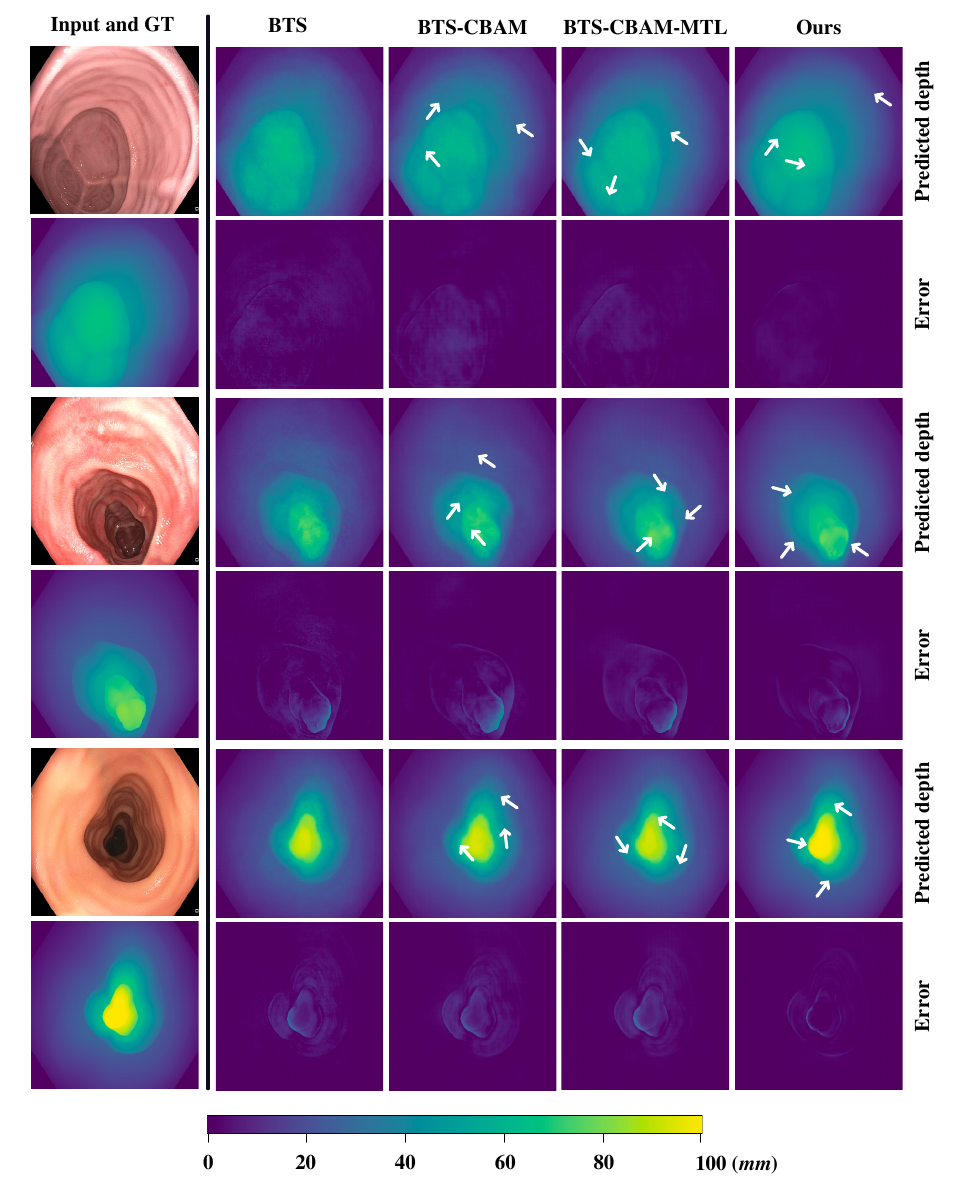}
    \caption{Qualitative comparison between our different network configurations. We show the input image, its corresponding ground truth depth maps, the depth predictions for each network configuration, and its corresponding absolute error maps. White arrows show the positive impact of each network configuration concerning the previous one. The addition of the CBAM modules partially reduces the smooth transitions of our baseline method. Leveraging our MTL approach reduces the absolute error at transition zones but does not recover an accurate estimation at areas with low texture. Explicitly enforcing consistency among tasks achieves an enhanced depth estimation.}
    \label{fig:qualitative_ntw_ablation}
\end{figure}

In Fig.~\ref{fig:qualitative_ntw_ablation}, we compare the absolute error maps from all the proposed network configurations of our ablation study to observe the effect of each approach. Lower absolute error maps are achieved after each incorporated module, emphasising their positive impact towards an enhanced depth estimation. Our baseline method inaccurately estimates the depth values at transition zones, e.g., on the folds of the colon and occluded regions. The addition of CBAM modules reduces the absolute depth error at areas corresponding to the folds of the colon but not at occlusion zones where depth variation can be higher. Incorporating our surface normal predictor diminishes the absolute error of our depth estimation at these zones. However, the highest absolute errors within each depth map remain in these areas. Our BTS-CBAM-MTL-X-TC approach addresses these cases by explicitly enforcing consistency among both predictions. Our proposed framework leads to an overall lower absolute error map but also achieves a sharper depth prediction with fewer visual artefacts.

\begin{figure}[!th] 
    \centering
    \includegraphics[width=\linewidth]{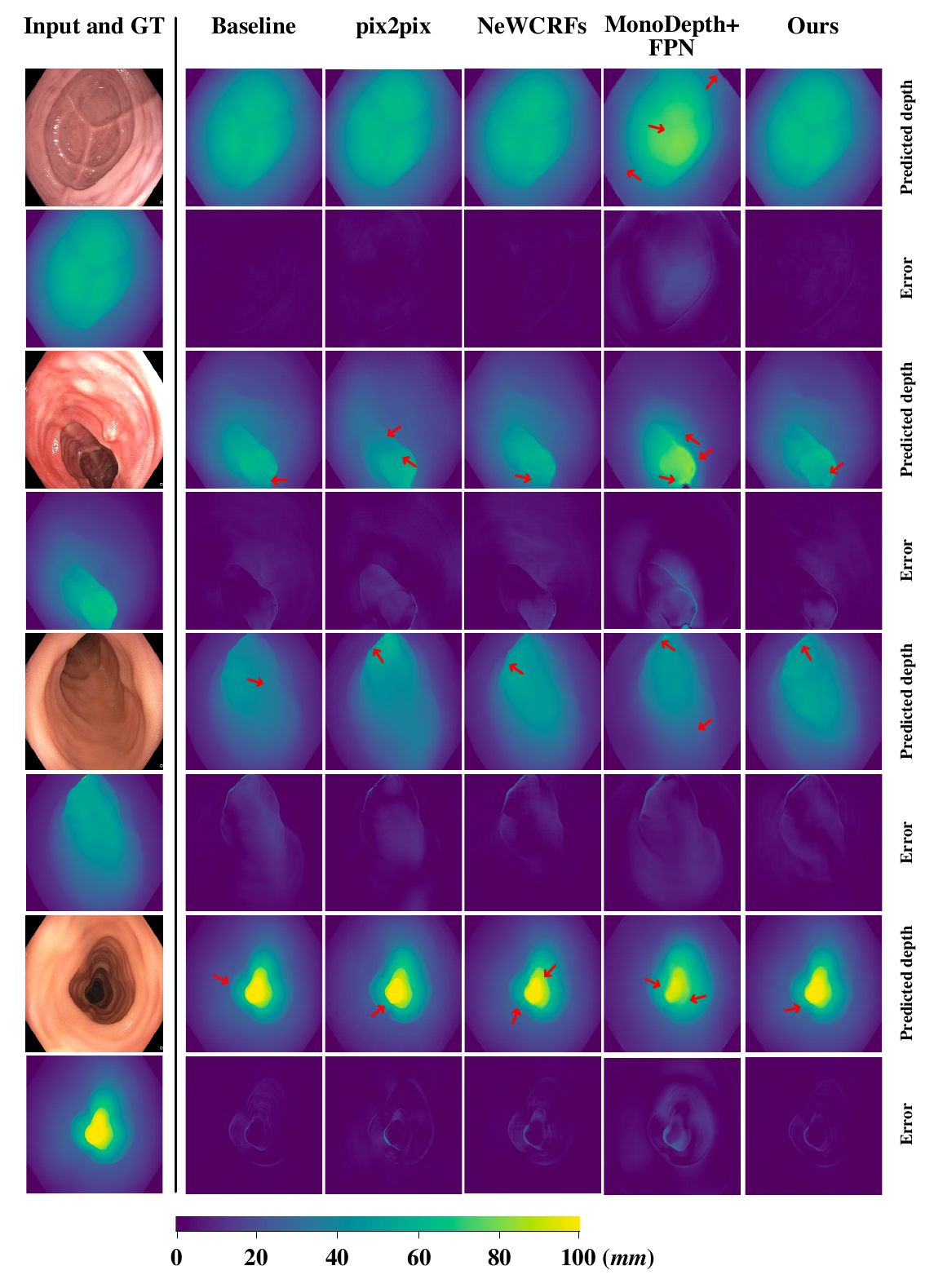}
    \caption{Qualitative comparison between all evaluated methods in this study. One sample from each colon segment and its corresponding ground truth depth map is given from the testing set. The depth prediction of each method is provided with its corresponding absolute error map. Red arrows specify challenging regions for each method, e.g., folds of the colon, polyps, occlusion zones, low-texture regions, and areas with low lighting conditions.}
    \label{fig:qualitative_all}
\end{figure}

Fig.~\ref{fig:qualitative_all} shows a qualitative comparison between all the state-of-the-art methods evaluated in this study: BTS, pix2pix, NeWCRFs, MonoDepth+FPN, and our Col3D-MTL framework. From the testing set, one sample from each colon segment with its corresponding ground truth and depth prediction is given. The highest absolute error maps are generated by MonoDepth+FPN, followed by pix2pix, which recovered an overall depth representation of the scene but not as sharp as our baseline method. Our baseline method improves pix2pix but generates smooth changes in transition zones. NeWCRFs surpasses our baseline method in regions with low texture but cannot accurately estimate depth in regions with high depth variability. Our proposed framework leads to the lowest absolute error maps among all the previous methods, addressing low texture and regions with increased depth variability.

%
\section{Discussion}
\label{sec:discussion}
Monocular depth estimation methods rely on the representation of visual cues to recover the depth information from a single image. Convolutional neural networks have shown an outstanding performance in extracting local feature representations but need more contextual information to relate them properly. 
Therefore, we explored the use of CBAM modules at each multi-scale stage of the decoder and skip connections of our baseline method to incorporate attention mechanisms and leverage global context awareness. Our experiments show an improved performance on all evaluation metrics (see top of Table~\ref{tab:MTL-results}) and a refined depth map with respect to our baseline (second and third columns of Fig.~\ref{fig:qualitative_ntw_ablation}). However, our BTS-CBAM configuration shows a tendency to create blurry regions. 
To further enhance the extraction of salient features and leverage the orientation of the scene, we integrate a geometrically related task, namely surface normal estimation, into our previous network with attention mechanisms. We have designed an independent surface normal decoder with novel unit normal computation blocks and incorporated it into our baseline method with CBAM modules. Our BTS-CBAM-MTL approach using an $L_{1}$ loss function to optimise the surface normal decoder yields the best trade-off performance considering all evaluation metrics for both tasks (see middle of Table~\ref{tab:MTL-results}). The extraction of salient geometrical features leads to a better scene representation, partially reducing the generation of blurry regions of our BTS-CBAM network (third and fourth columns of Fig.~\ref{fig:qualitative_ntw_ablation}). However, without an explicit constraint to enforce cross-task consistency, we do not achieve a refined depth prediction at regions with high-depth variability, such as the folds of the gastrointestinal tract and occlusion zones.
To explicitly enforce consistency among both predictions, we implement a cross-task consistency scheme. In order to evaluate consistency among depth and surface normal predictions, we implement a warping module based on DIG to generate a warping surface normal map from our depth estimation. Our cross-task consistency scheme aims to minimise the RMSE between the warped surface normal and the surface normal prediction of our decoder. We define our final loss function as a weighted sum of each loss function that optimises our depth estimation, surface normal prediction, and cross-task consistency module. The results of our ablation study show that the best set of $\lambda$ weighting factors consists of $\lambda_{1} = 0.5$, $\lambda_{2} = 0.3$, and $\lambda_{3} = 0.2$ (Table~\ref{tab:MTL-XTC-results}).
Our proposed Col3D-MTL framework surpasses all previous approaches in our network configuration ablation study (see bottom of Table~\ref{tab:MTL-results}). We can observe that our approach leads to the most refined depth maps with the lowest absolute error maps (fifth column of Fig.~\ref{fig:qualitative_ntw_ablation}). We can notice an improved accuracy on transition zones, e.g., on the folds of the colon or in regions containing polyps. Furthermore, the enhanced, detailed regions in our depth estimation show consistency with the accurate surface normal predictions (see Fig.~\ref{fig:qualitative_cases}).

We set a new benchmark on the C3VD dataset, in which our proposed method achieves the best performance among the evaluated state-of-the-art methods on monocular depth estimation (Table~\ref{tab:comparative-results}). Our approach yields a relative improvement of 3.2\% on SILog, 12.6\% on RMSE, and 10.4\% on $\delta_{1}$ over our baseline. We further validate our approach by evaluating each method on each colon segment separately. On all colon segments, except for the caecum, our proposal achieves the best $\delta_{1}$ value. However, the most remarkable achievement only considers the unseen descending colon segment, in which we surpass all the networks presented in this study (Table~\ref{tab:per-segment}). By leveraging local feature representation and global context awareness with cross-task consistency among our two geometrically related tasks, we refine our depth predictions, which demonstrate lower absolute error maps than the other approaches (see Fig.~\ref{fig:qualitative_all}).

\section{Conclusion}
\label{sec:conclusion}
Colonoscopy screening remains the gold standard for diagnosing and treating inflammatory bowel diseases. However, due to its challenging anatomical environment and variable conditions, it is a highly operator-dependent procedure, which usually leads to a high missed-detection rate. Although several approaches have been proposed to detect and segment instruments and polyps, recovering the 3D scene information to perform a quantitative assessment has not been widely studied. Recovering the depth information of a scene is the first step in a 3D reconstruction pipeline. 
We have identified the current challenges of monocular depth estimation methods and developed our proposed framework towards its applicability in the colonoscopy domain. We selected BTS as our baseline monocular depth estimation method. Given the outstanding local feature representation of convolutional neural networks, our proposed method leverages CBAM attention mechanisms to improve global context awareness and to relate our extracted local features, a surface normal decoder with novel unit normal computation blocks to enhance the 3D representation of the scene, and a cross-task consistency scheme to explicitly enforce consistency among depth and surface normal predictions. To demonstrate the impact of each module, we have provided a comprehensive experimental setup which validates our Col3D-MTL network. 
Our framework is compared against other state-of-the-art monocular depth estimation methods on the C3VD dataset, which is entirely recorded with a high-definition clinical colonoscope on a silicone phantom model that mimics the vascular patterns and the specular appearance of the colon mucosa.
Our quantitative results show that our proposed network outperforms current state-of-the-art methods. The most remarkable improvement of our method is achieved on a colon segment that was not given during the training stage. Our qualitative results support adding each proposed module towards a refined feature representation of the colon scene. 

\subsection*{Limitations and future work}
\noindent The limitations of the proposed framework include inaccurate depth and surface normal predictions in regions with low lighting conditions, usually located at the farthest region of the colonoscopy scene. Other cases in which our surface normal decoder does not recover the surface orientation include small regions with high orientation variability, usually encountered as the region gets farther from the colonoscope.
Given the need for properly annotated real-world colonoscopy data for 3D reconstruction, our future work will focus on improving the robustness of our method to translate our method from synthetic datasets to real-world clinical data.

\section*{Declaration of Competing Interest}
\noindent The authors declare that they have no known competing financial interests or personal relationships that could have appeared to influence the work reported in this paper.

\section*{CRediT authorship contribution statement}
\noindent \textbf{Pedro Chavarrias:} Conceptualisation, Methodology, Investigation, Data curation, Software, Validation, Formal analysis, Writing - original draft, Writing - review \& editing. \textbf{Andrew Bulpitt:} Writing - review \& editing and Supervision. \textbf{Venkat Subramanian:} Writing - review \& editing and Supervision. \textbf{Sharib Ali:} Conceptualisation, Methodology, Investigation, Software, Formal analysis, Writing - original draft, Writing - review \& editing, Primary Supervision and Project administration.

\section*{Acknowledgements}
\noindent This work was undertaken on ARC4, part of the High Performance Computing facilities at the University of Leeds, UK.



\bibliographystyle{model2-names.bst}\biboptions{authoryear}
\bibliography{mybibfile}

\end{document}